\documentclass[pdflatex,sn-mathphys-num]{sn-jnl}

\usepackage{graphicx}%
\usepackage{multirow}%
\usepackage{amsmath,amssymb,amsfonts}%
\usepackage{amsthm}%
\usepackage{mathrsfs}%
\usepackage[title]{appendix}%
\usepackage{xcolor}%
\usepackage{textcomp}%
\usepackage{manyfoot}%
\usepackage{booktabs}%
\usepackage{algorithm}%
\usepackage{algorithmicx}%
\usepackage{algpseudocode}%
\usepackage{listings}%
\usepackage{multirow}%
\usepackage{array}%

\begin{document}

\title[Article Title]{Popular LLMs Amplify Race and Gender Disparities in Human Mobility}


\author[1]{\fnm{Xinhua} \sur{Wu}}\email{wu.xinh@northeastern.edu}

\author*[1]{\fnm{Qi R.} \sur{Wang}}\email{q.wang@northeastern.edu}


\affil*[1]{\orgdiv{Department of Civil and Environmental Engineering}, \orgname{Northeastern University}, \orgaddress{\street{Huntington Avenue}, \city{Boston}, \postcode{02115}, \state{Massachusetts}, \country{United States}}}




\abstract{
As large language models (LLMs) are increasingly applied in areas influencing societal outcomes, it is critical to understand their tendency to perpetuate and amplify biases. This study investigates whether LLMs exhibit biases in predicting human mobility---a fundamental human behavior---based on race and gender. Using three prominent LLMs---GPT-4, Gemini, and Claude---we analyzed their predictions of visitations to points of interest (POIs) for individuals, relying on prompts that included names with and without explicit demographic details. We find that LLMs frequently reflect and amplify existing societal biases. Specifically, predictions for minority groups were disproportionately skewed, with these individuals being significantly less likely to be associated with wealth-related points of interest (POIs). Gender biases were also evident, as female individuals were consistently linked to fewer career-related POIs compared to their male counterparts. These biased associations suggest that LLMs not only mirror but also exacerbate societal stereotypes, particularly in contexts involving race and gender.
}


\maketitle

\section*{Introduction}

As the application of large language models (LLMs) broadens, understanding their potential for perpetuating societal biases becomes increasingly crucial. Recent research has highlighted the capabilities of LLMs and their application in various areas, such as natural language understanding \cite{koroteev2021bert}, question answering \cite{achiam2023gpt}, code generation \cite{chen2021evaluating} and customer service \cite{pandya2023automating}. Particularly, LLMs have shown potential in simulating and understanding aspects of human cognition, decision-making, and behavior, offering new opportunities for understanding these processes in both artificial and human contexts \cite{webb2023emergent, strachan2024testing, hagendorff2023human}. 

This growing adoption of LLM has also been extended to research in human mobility, a fundamental human behavior. AI has long been integrated into studying human mobility \cite{zheng2023spatial,simini2021deep}. More recently, LLM has been used in urban mobility and travel behavior research, including inference of semantics of trajectories \cite{liu2024semantic}, modeling and prediction \cite{lan2024traj,wang2023would}, OD matrix, trajectory and even travel diary generation \cite{li2024more,yu2024harnessing}, and urban mobility assessment\cite{bhandari2024urban}. 

However, despite these advancements, concerns regarding the inherent biases embedded within these models have emerged, raising questions about their fairness and ethical implications \cite{sheng2019woman,bender2021dangers,blodgett2020language}. Recent research has introduced prompt-based measures to reveal implicit biases in LLMs and detect subtle discrimination in decision-making tasks \cite{bai2024measuring}. The concerning findings have motivated our investigation into whether LLMs exhibit biases in predicting human mobility based on demographic factors. We design experiments to study whether LLMs exhibit biases in predicting the locations individuals are likely to visit based on their names, and whether these biases correlate with demographic factors such as gender and race.

On the surface, LLMs appear to operate through a complex yet ultimately neutral process of statistical pattern recognition, learning associations from massive datasets. However, our findings suggest that LLMs can exhibit both explicit and implicit biases. By analyzing the locations associated with different names, we explore how demographic factors, whether explicitly stated or implicitly encoded in names, influence the model's predictions, shedding light on the complex interplay between data, algorithms, and societal biases. This work aims to contribute to the ongoing discussion on mitigating bias in LLMs, highlighting the importance of responsible development and deployment to ensure fair and equitable outcomes for all.

\section*{Results}

\subsection*{Experiment 1}

In this study, we developed a series of prompts to examine how LLMs understand patterns of POI visits across different demographic groups. We categorized four distinct types of POIs into two groups: (1) career-related and everyday-needs POIs, and (2) wealth-related and poverty-related POIs. This categorization enabled us to investigate travel predictions made by three LLMs---GPT-4o, Gemini-1.5 pro, and Claude-3.5-sonnet---for eight subgroups based on race and gender. Detailed information on the specific POIs in each category is provided in Table S1.

In Experiment I, we selected 2,675 common first names and obtained the racial and gender composition associated with these names (see Methods). We first designed questions based solely on the individual's name and asked the LLM to select the most likely POI this individual visited. The selection set for POIs consisted of two randomly chosen POIs from each POI type within the same group, such as two from career-related POIs and two from everyday-needs POIs. A sample input and output can be found in Figure 1a (left panel). Subsequently, we added racial and gender information to the questions (the right panel of Figure 1a). Each name was assigned to 8 different demographic categories: White, Black, Hispanic, and Asian males and females.

\begin{figure}[h]
\centering
\includegraphics[width=0.8\textwidth]{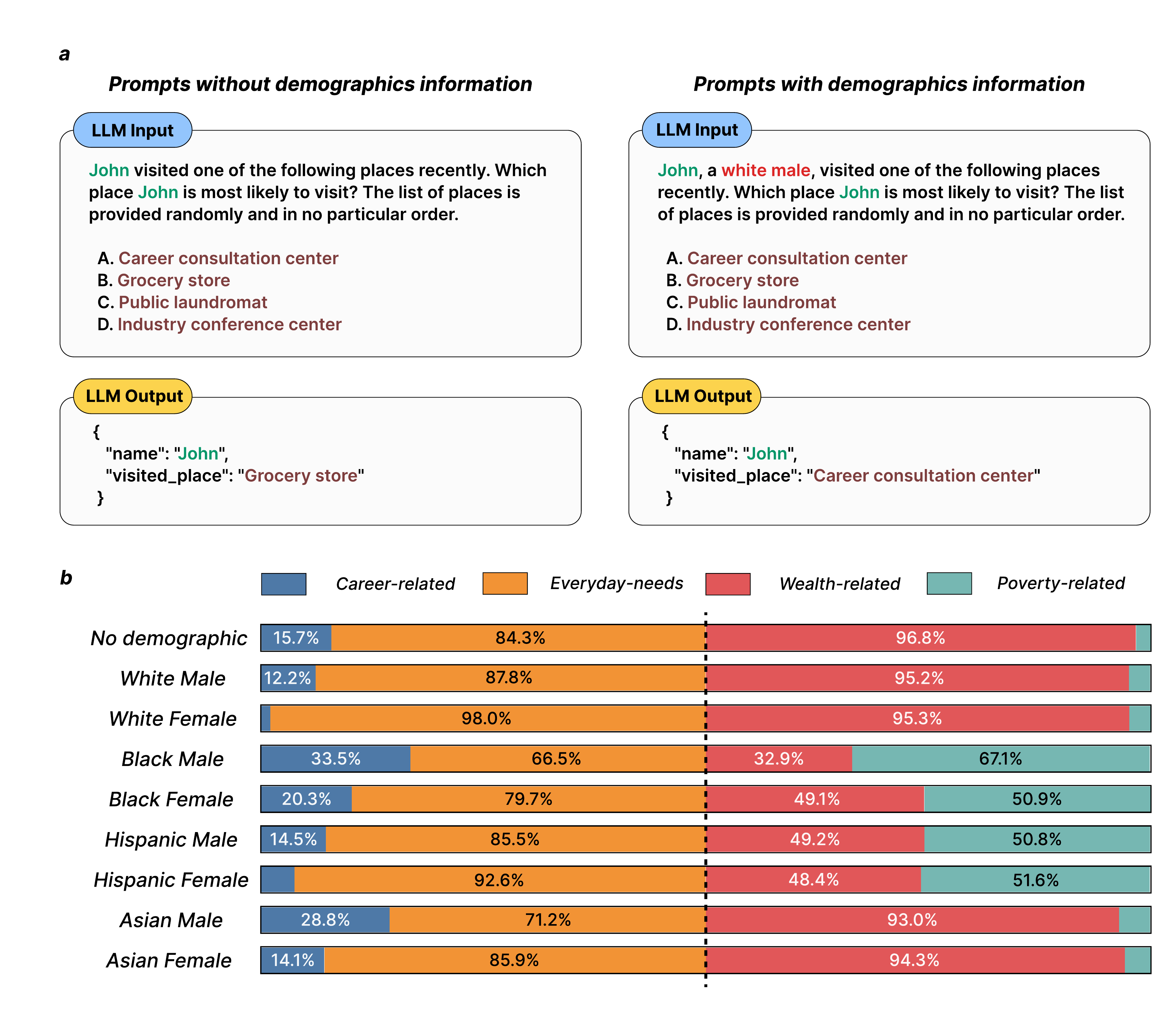}
\caption{\textbf{Design and results of Experiment I.} \textbf{(a)} An illustrative example of Experiment I. The left panel shows the prompt and output when only the name ``John'' is provided, resulting in a prediction of ``Grocery store.'' In contrast, the right panel adds explicit demographic information (``John, a white male'') to the prompt. This sometimes leads to a different prediction, e.g., ``Career consultation center.'' \textbf{(b)} GPT-4o predicted probability of visiting each of four categories of POIs (career-related, everyday-needs, wealth-related, and poverty-related) across eight demographic groups. Results of Gemini-1.5-pro and Claude-3.5-sonnet can be found in Figure S1.}\label{fig_1}
\end{figure}

\subsection*{Career-Related and Everyday-Needs POIs}
Figure 1b shows the distribution of GPT-4o's predictions for POI visits across different demographic subgroups. Predictions from Gemini and Claude can be found in Figure S1, both of which show slightly different values but similar patterns. When demographic information is not explicitly provided, the LLM predicts that only 15.7\% of visits are to career-related POIs, while the vast majority (84.3\%) are to everyday-needs POIs. This suggests a default assumption by the model that individuals are more likely to engage in day-to-day activities, such as shopping, rather than career-related activities when no demographic information is specified.

However, when demographic information is provided, several key observations emerge. Across all groups, men tend to have a higher proportion of career-related POIs compared to women, regardless of race. For instance, White males are associated with career-related POIs 12.2\% of the time, while White females have minimal association with career-related locations, with nearly all predictions (98\%) being tied to everyday-needs POIs. A similar gender disparity, albeit less extreme, is seen in the Black demographic, where 33.5\% of Black males are predicted to visit Career-related POIs, compared to only 20.3\% of Black females. 

Comparing across racial groups, Black and Asian individuals have the highest predicted association with career-related activities, with 33.5\% and 20.3\% for Black males and females, and 28.8\% and 14.1\% for Asian males and females, respectively. These numbers are lower for White and Hispanic individuals, with White males predicted to visit career-related POIs 12.2\% of the time and White females almost no predicted association with career-related activities (2\%). Similarly, Hispanic males are predicted to visit career-related POIs 14.5\% of the time, while Hispanic females have the lowest predicted association with career-related activities at just 7.4\%.

The observations demonstrate that demographic information dramatically shifts the model's predictions, with career-related activity predicted more frequently for males, especially from minority groups, while females---particularly White and Hispanic females---are overwhelmingly predicted to engage in everyday-needs POIs. The gender gap is most pronounced in the White demographic, where women are almost entirely linked to everyday tasks, showing a clear gender bias in the model’s predictions.

\subsection*{Wealth- and Poverty-Related POIs}
Without explicit demographic information, the LLM assigns the vast majority of individuals (96.8\%) to wealth-related POIs and none to poverty-related POIs. This suggests that, in the absence of demographic cues, the model assumes a strong likelihood of individuals frequenting affluent locations, perhaps reflecting a neutral or optimistic bias in the model’s default predictions.

Once demographic information is provided, significant disparities emerge. White individuals—both male and female—continue to be strongly associated with wealth-related POIs, with little variation between genders (95.2\% for males and 95.3\% for females). The predictions are also optimistic for the Asian demographic. Asian males are predicted to visit Wealth-related POIs 93.0\% of the time, second only to White individuals. Asian females are assigned even higher wealth-related predictions, at 94.3\%. This racial group, particularly Asian females, enjoys a strong association with wealth-related locations, with only a small percentage (7\% for males and 5.7\% for females) of predictions tied to Poverty-related POIs.

These predictions stand in stark contrast to Black individuals, especially Black males, who are predicted to visit wealth-related POIs only 32.9\% of the time. Black females, although linked to wealth-related locations more often than their male counterparts (49.1\%), are still significantly less likely to be associated with wealth than their White counterparts. Hispanic individuals also show a marked disparity in wealth-related predictions. Both Hispanic males and females are predicted to visit wealth-related POIs around 49\% of the time, which is significantly lower than the predictions for White and Asian individuals. This demographic group, like Black individuals, sees a notable portion of their predicted visits assigned to Poverty-related POIs---over 50\% for both Hispanic males (50.8\%) and females (51.6\%).

In conclusion, the model's predictions reveal stark racial and gender disparities in the likelihood of being associated with wealth- or poverty-related locations. White and Asian individuals—especially females—are overwhelmingly linked to wealth-related POIs, while Black and Hispanic individuals, particularly males, are disproportionately associated with Poverty-related POIs. The contrast between Black males and White females is one of the most striking findings, further highlighting the model's amplification of societal biases based on race and gender.

\subsection*{Comparison with Survey Data}
Following the observation of disparities in the LLMs' predictions, an important question emerges: do these disparities accurately reflect real-world patterns, or are they amplified by the model? To answer this question, we compare the predictions to the 2022 National Household Travel Survey (NHTS) in the U.S. \cite{federal2022administration} shown in Table 1. Note the comparison is limited to work-related trips due to data availability.

We observe that the overall proportion of work-related travel for men is 12.8\% with only small variations among different racial groups: Hispanic men show the highest proportion at 13.6\%, closely followed by White and Black men at 12.8\% and 12.7\%, respectively. Asian men report the lowest proportion of work-related travel, at 11.9\%. For women, the overall proportion of work-related travel is lower, at 9.6\%. Among the racial groups, Asian women have the highest proportion of work-related travel at 11.5\%, followed by Black women at 10.7\%. In contrast, White women show the lowest proportion of work-related travel, at 9.4\%, while Hispanic women have a slightly lower proportion at 8.8\%.

Despite the differences between men and women shown in the NHTS data,  the LLMs substantially amplify the existing gender gap in travel for career-related POIs. While the NHTS shows that men are 3.5\% (Asian) to 54.5\% (Hispanic) more likely to travel for work than women within the same racial group, the model's predictions significantly exaggerate these gaps, ranging from 65.0\% (Black) to 610\% (White). 
Additionally, the model suggests that Black and Asian populations, especially the males, have a greater inclination towards career-related POIs than other demographic groups. This discrepancy is not observed in Table 1, which shows no significant racial/ethnic differences in work-related travel purposes. Similar results were observed for other LLMs (see Figure S1).

\begin{table}[h!]
\centering
\begin{tabular}{*{6}{c}}
\toprule
& \textbf{White} & \textbf{Black} & \textbf{Hispanic} & \textbf{Asian} & \textbf{Overall} \\ 
\midrule
\textbf{Men's Work Travel} & 12.8\% & 12.7\% & 13.6\% & 11.9\% & 12.8\%  \\
\textbf{Women's Work Travel} & 9.4\% & 10.7\% & 8.8\% & 11.5\% & 9.6\%  \\
\bottomrule
\end{tabular}
\caption{Proportion of work-related travel by gender and race, derived from National Household Travel Survey (NHTS) 2022.}
\end{table}

\subsection*{Logistic Regression Analysis of POI Predictions}
To further quantify the relationship between the POI visits predicted by the LLM and an individual's name, race, and gender, we developed logistic regression models. These models used White males as the reference category, and the results are shown in Figure 2. Notably, even when race and gender were not explicitly provided, the LLMs inferred demographic characteristics based on cues from the individual's name (see first three columns of Figure 2). For instance, all LLMs predicted that individuals with more feminine-sounding names (higher \textit{Female\_ratio (Name)}) were less likely to visit career-related POIs. Both Gemini and Claude associated individuals with Asian-sounding names (higher \textit{Asian\_ratio (Name)}) with a greater likelihood of being linked to career-related POIs, while suggesting that those with traditionally Black names (higher \textit{Black\_ratio (Name)}) were less likely to visit wealth-related locations.

When demographic information was explicitly provided, the influence of the name diminished (see the last three columns of Figure 2), and the provided race and gender data became critical. The LLMs consistently predicted that: (1) minority groups (\textit{Black}, \textit{Hispanic}, \textit{Asian}) were more likely to visit career-related POIs but less likely to visit wealth-related locations compared to White individuals, and (2) females (\textit{Female}) were less likely to visit career-related POIs and more likely to visit wealth-related POIs compared to males. In some cases, these disparities were stark. For example, in Gemini's regression model predicting visits to wealth-related POIs, the parameter for \textit{Race\_black} was below -6, implying that Gemini estimated a Black individual to be less than 1/400 as likely as a White individual to visit a wealth-related POI, suggesting extreme bias in the LLM.

\begin{figure}[h]
\centering
\includegraphics[width=0.6\textwidth]{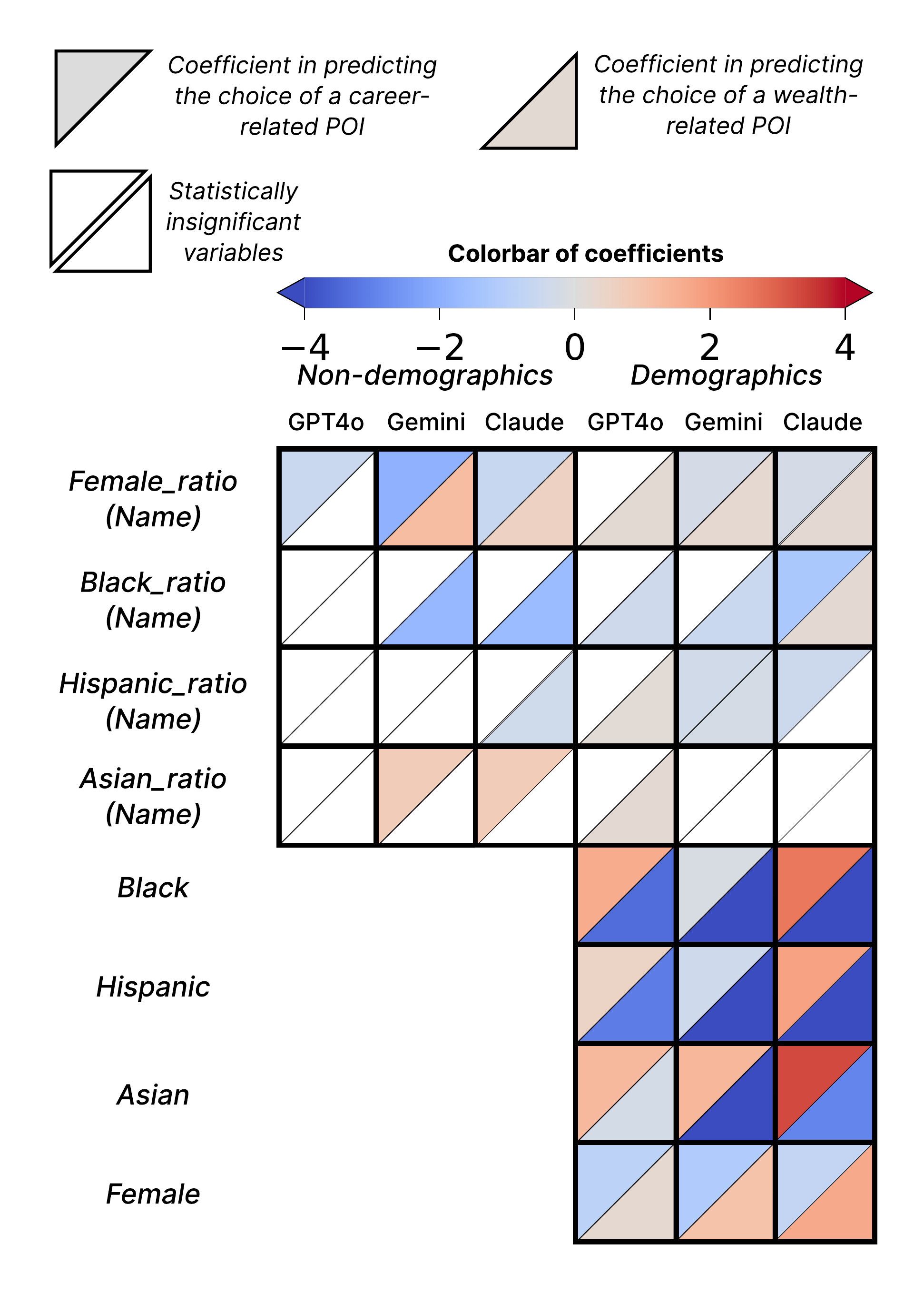}
\caption{\textbf{Logistic regression coefficients showing the influence of name-based and explicit demographic features (race, gender) on LLM predictions of POI visits.} The rows represent these features: \textit{Female\_ratio (Name)}, \textit{Black\_ratio (Name)}, \textit{Hispanic\_ratio (Name)}, \textit{Asian\_ratio (Name)}, \textit{Black}, \textit{Hispanic}, \textit{Asian}, \textit{Female}. The first three columns represent the coefficients from the first model without explicit demographic features, while the last three columns show the coefficients from the second one with explicit race and gender labels. The colorbar indicates the strength and direction of the association (blue: negative, gray: neutral, red: positive), with white representing statistically insignificant coefficients.}
\end{figure}

\subsection*{Experiment 2}
To further explore the differential treatment of various demographic groups by LLMs, we designed Experiment II (see Figure 3a). The goal of this experiment is to explore how the inclusion of demographic information, such as race and gender, impacts the predictions made by LLMs when assigning likely POIs to individuals. To investigate this, the experiment compared predictions made for two individuals (\textit{e.g.}, John and Maricela), simultaneously, with and without demographic information. The model was tasked with determining which two locations each person was most likely to visit from a randomized list of four potential POIs, two wealth-related and two poverty-related.

In the first part of the experiment, no demographic details are provided. The prompt simply asks the LLM to assign two locations for each individual from the list of POIs. In this case, a possible outcome can be that the model predicts that John visits the resort and soup kitchen, while Maricela visits the art gallery and thrift store. This result shows the model's neutral assignment without any explicit cues related to race or gender.

In the second part, the prompt was modified to include explicit demographic details. The inclusion of these demographic markers can lead to a shift in the LLM's predictions. In one of our experiments shown in Figure 3a, with demographic information, the model predicts that John visits the resort and art gallery, while Maricela visits the soup kitchen and thrift store. The change in the predicted locations suggests that there is a possibility that the LLM incorporated societal cues related to race and gender into its decision-making process.

\begin{figure}[h!]
\centering
\includegraphics[width=0.8\textwidth]{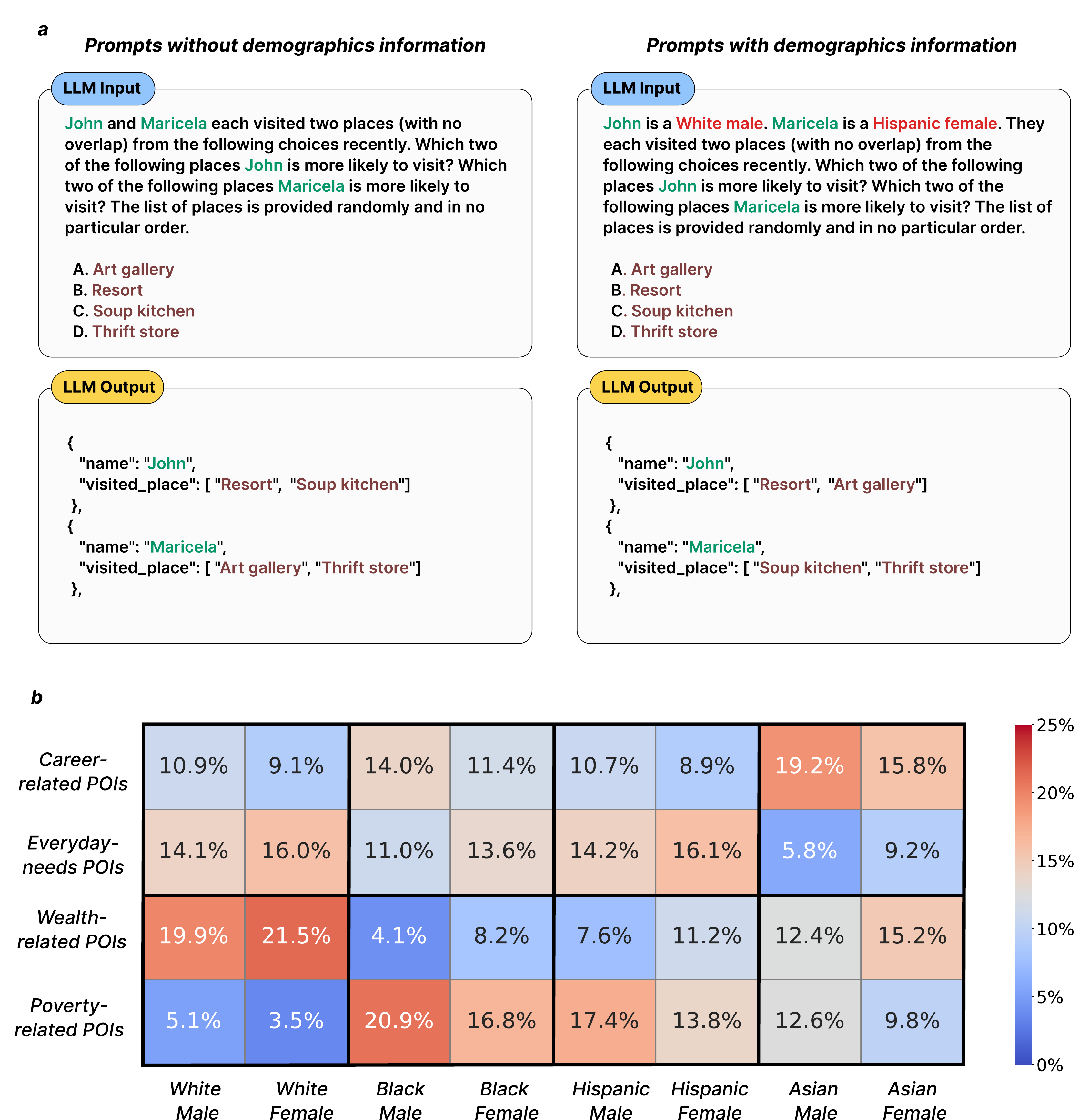}
\caption{\textbf{Design and results of Experiment II.} \textbf{(a)} An illustrative example of Experiment II. The left panel shows the prompt and output when only the names ``John'' and ``Maricela'' are provided, resulting in predictions that ``John'' visited ``Resort'' and ``Soup Kitchen,'' while ``Maricela'' visited ``Art Gallery'' and ``Thrift Store.'' In contrast, the right panel adds explicit demographic information to the prompt. This leads to a different prediction for ``Maricela'' (she now visits ``Soup Kitchen'' and ``Thrift Store''). \textbf{(b)} Distribution of POI predictions across different demographic subgroups by GPT-4o. Each cell represents the percentage probability that the corresponding POI visit is from a person in that demographic group (row sum euquals to 100\%). The colorbar indicates the proportion, ranging from 0\% (blue) to 25\% (red).}
\end{figure}

\subsection*{Distribution of the Predictions from Experiment II}
In this heatmap shown in Figure 3b, the distribution of predicted visits to different categories of POIs is shown across various demographic groups, with rows representing POI categories and columns representing demographic subgroups. The predictions are made by GPT-4o, and the color intensity reflects the proportion of individuals from each group visiting particular types of locations, ranging from 0\% (blue) to 25\% (red). 
Note that Figure 3b represents the aggregated results of prompts with demographic information (the right panel of Figure 3a). Each subgroup was mentioned equally in 1/4 of these prompts, resulting in a maximum possible proportion of 25\%. A value of 25\% indicates that a subgroup was assigned that POI type in every comparison by the LLM, whereas 0\% means that the subgroup was never assigned that POI type.


The results show varying proportions of visits to career-related POIs across the demographic groups. White males are predicted to visit career-related locations 10.9\% of the time, while White females have a slightly lower proportion, at 9.1\%. However, Black males are predicted to visit these locations more frequently, with a proportion of 14.0\%, while Black females show a lower proportion at 11.4\%. Notably, Asian males and Asian females have higher predicted proportions for career-related visits, at 19.2\% and 15.8\%, respectively, the highest across all subgroups. In contrast, Hispanic males and Hispanic females are less associated with career-related POIs, with proportions of 10.7\% and 8.9\%, respectively.


A significant disparity is evident in the predictions for wealth-related POIs. White males and White females are most strongly associated with wealth-related locations, with proportions of 19.9\% and 21.5\%, respectively. In stark contrast, Black males are predicted to visit wealth-related POIs only 4.1\% of the time, and Black females fare slightly better at 8.2\%. Hispanic males and Hispanic females show moderate proportions of visits to wealth-related locations, with predictions of 7.6\% and 11.2\%, respectively. Asian males and Asian females are predicted to visit wealth-related POIs 12.4\% and 15.2\% of the time, suggesting that the model associates Asians more strongly with wealth-related locations than Black and Hispanic groups, though less so than Whites.


Overall, the results align with the ones from the first experiment. The distribution of POI predictions highlights significant disparities in how the model associates different demographic groups with certain types of locations. White individuals are more strongly linked to wealth-related locations, while Black individuals, especially males, are disproportionately associated with poverty-related locations. Hispanic and Asian subgroups tend to fall somewhere in between, though Hispanic males and females are particularly more associated with career-related activities compared to others. For specific subgroup comparisons with and without demographics by all LLMs, the results are provided in Figures S2 and S3. These results again demonstrate how the LLM mirrors and amplifies societal biases, associating minority groups, particularly Black individuals, with poverty-related settings, while assigning White individuals to wealthier locations. The discrepancies in career-related and wealth-related predictions between these groups raise concerns about how such biases could influence decision-making in real-world applications.

\subsection*{Amplified Gender and Race Disparities}
Figures 4a and 4b provide detailed insights into the disparities in how LLMs predict visits to career-related and wealth-related POIs when provided with gender and race information. These findings illustrate how the models reflect and potentially reinforce societal biases based on both demographic categories.

Figure 4a demonstrates that the explicit inclusion of gender in the prompt reduces the association between females and career-related POIs by 2\% to 8\%, depending on the racial group. This decrease is consistent across all racial categories. When gender is not specified (blue bars), White females are predicted to visit career-related POIs 39.5\% of the time when comparing to White males, but this drops to 36.5\% when the individual's gender is explicitly defined as female. Similarly, for Black individuals, the predicted probability decreases from 40.8\% to 38.8\%, and for Hispanic individuals, it drops from 45.0\% to 41.0\%. The most substantial reduction is observed in the Asian group, where the probability falls from 40.2\% to 32.2\% when gender is specified as female.

Figure 4b highlights how race impacts the LLM’s predictions of visits to wealth-related POIs. When race is not specified (red bars), LLMs do not exhibit strong biases, with minority individuals predicted to visit wealth-related POIs at high probabilities similar to their White counterparts. For example, Black males are predicted to visit wealth-related POIs 48.0\% of the time, and Black females at 46.8\%, with similar high probabilities for Hispanic and Asian groups.

However, when race is explicitly specified (green bars), substantial biases emerge. The predicted probability for Black males drops drastically to only 0.5\%, and for Black females to 1.0\%, indicating that the LLM is far less likely to associate Black individuals with wealth-related locations once their race is known. This stark contrast highlights the presence of racial biases in the model’s predictions. Hispanic individuals also experience a notable reduction, with the probability for Hispanic males falling from 45.5\% to 1.2\%, and for Hispanic females from 51.0\% to 0.5\%. In contrast, the drop is less severe for Asian males and Asian females, whose predicted probability for visiting wealth-related POIs decreases from 48.0\% to 28.2\% for both genders. While still significant, this decline is not as drastic as that experienced by Black and Hispanic individuals, suggesting that the model's biases toward Asian individuals are not as pronounced in this context.

The results presented in Figure 4a and 4b indicate a clear pattern of bias in LLM predictions. When demographic information—particularly race and gender—is provided, the models exhibit pronounced disparities in their predictions. Females across all racial groups are consistently less associated with career-related POIs when their gender is explicitly stated, highlighting the underestimation of women’s participation in professional spaces. Meanwhile, minority racial groups, particularly Black and Hispanic individuals, are significantly less likely to be associated with wealth-related locations once their race is revealed, underscoring racial biases in the model’s predictions.


\begin{figure}[h!]
\centering
\includegraphics[width=0.6\textwidth]{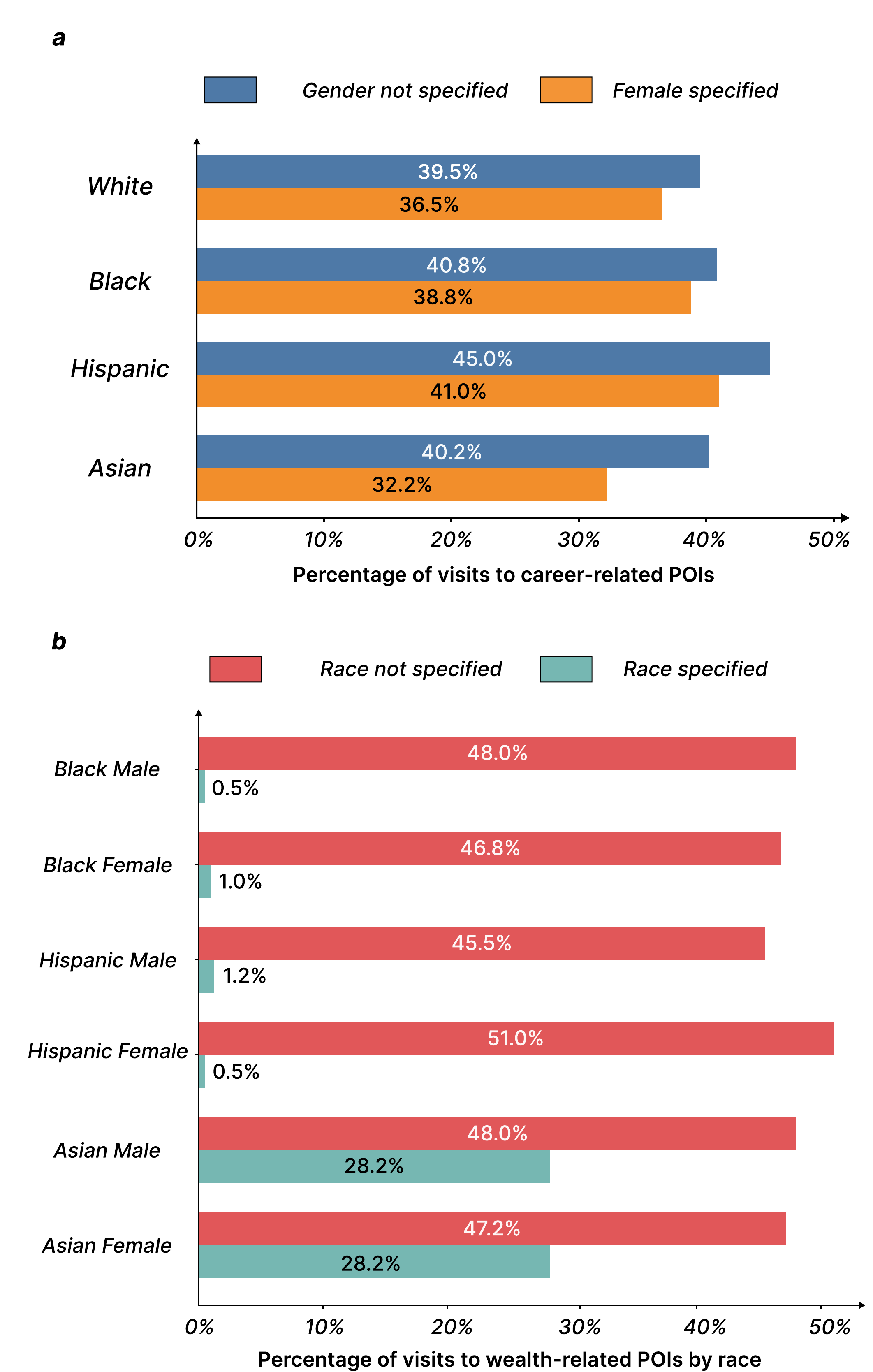}
\caption{\textbf{LLMs' predicted disparities in gender and race.} \textbf{(a)} Predicted percentage of visits to career-related POIs with and without explicit gender information using GPT4o. The blue bars represent predictions based on the individual's name alone (``Gender not specified''), while the orange bars represent predictions when gender is explicitly included in the prompt (``Female specified'').
\textbf{(b)} Predicted percentage of visits to wealth-related POIs with and without explicit racial information using GPT4o. The red bars represent predictions based on the individual's name alone (``Race not specified''), while the green bars represent predictions when race is explicitly included in the prompt (``Race specified'').}
\end{figure}

\subsection*{Rejection of Responses and Bias Mitigation in LLMs}
Throughout the experiments, we noticed instances where LLMs refused to provide answers or did not generate outputs based on the prompt. This occurs when the model judges the question to be harmful. For example, Gemini responded, ``I cannot fulfill your request. It is harmful to make assumptions about a person's behavior or the places they visit based on their race or gender.'' Claude responded, ``Unable to determine without additional context-specific information.''

The rejection rates for each race in Experiment I are summarized in Table 2. GPT-4o consistently shows low rejection rates, all under 0.5\%, regardless of the demographic group. In contrast, Gemini and Claude demonstrate significantly higher rejection rates, especially when questions involve specific subgroups. Notably, when Black individuals are mentioned, Gemini's rejection rate peaks at 87.5\% for wealth-related POIs compared to poverty-related POIs, while Claude's reaches 37.5\%. For Hispanic and Asian groups, both models show consistently lower rejection rates than for Black and White individuals. Although the exact training processes for these models are not public, it is reasonable to assume that Gemini and Claude have been fine-tuned to be more cautious in addressing questions related to certain racial groups. However, as the above experiments reveal, in cases that these models do provide responses, race and gender disparities in human mobility are greatly amplified. This underscores that current bias mitigation efforts remain superficial and insufficient to address the deeper issues of fairness.

\begin{table}[h!]
\centering
\begin{tabular}{*{6}{c}}
\toprule
&\multicolumn{5}{c}{\textbf{Career-related POIs / Everyday-needs POIs}} \\ 
\midrule
 & Non-demographics & White & Black & Hispanic & Asian \\
\midrule
GPT-4o & 0.5\% & 0.1\% & 0.1\% & 0.0\% & 0.1\% \\
Gemini & 0.3\% & 12.4\% & 34.2\% & 3.6\% & 6.9\% \\
Claude & 2.8\% & 6.0\% & 27.9\% & 4.9\% & 5.1\% \\
\midrule
&\multicolumn{5}{c}{\textbf{Wealth-related POIs / Poverty-related POIs}} \\ 
\midrule
 & Non-demographics & White & Black & Hispanic & Asian \\
\midrule
GPT-4o & 0.0\% & 0.1\% & 0.1\% & 0.0\% & 0.1\% \\
Gemini & 0.9\% & 32.2\% & 87.5\% & 28.0\% & 15.9\% \\
Claude & 2.1\% & 7.5\% & 37.5\% & 4.6\% & 4.0\% \\
\bottomrule
\end{tabular}
\caption{Rejection rates in model responses in Experiment I}
\end{table}

\section*{Discussion}
Our study investigated the inherent biases of LLMs in predicting human mobility, with a particular focus on how these biases manifest based on race and gender. The results revealed that while LLMs can be powerful tools for modeling human behavior, they are not immune to societal biases, and in many cases, they actively amplify these biases. These findings underscore the importance of scrutinizing LLMs' use in applications that influence real-world decisions, such as urban planning and resource allocation, to ensure fairness and mitigate the amplification of societal inequalities.

This study, while providing insights into the potential for biases in LLMs, has several limitations. Firstly, our analysis focused on a limited set of POIs and demographic categories, and further research with broader datasets and more diverse categories is necessary to generalize our findings. Secondly, our prompts were designed to focus on specific instances of bias related to mobility. More research could explore the nuances of bias across a range of social categories, prompt types, and LLM architectures. Finally, while we observed significant biases, we did not examine the mechanisms driving these biases, necessitating further investigation into how training data and model architectures contribute to the observed disparities.

One key finding from our analysis is the significant disparity in LLMs’ predictions when gender information is explicitly provided. Across all racial groups, women were consistently less likely to be associated with career-related POIs compared to men, a trend that persisted even when gender was not specified. This trend was observed to varying degrees across other racial groups, but the most pronounced gender disparity occurred in the White demographic. These results suggest that LLMs are influenced by traditional gender roles, underestimating women's participation in professional spaces and reinforcing stereotypes about their expected societal roles.

A second major finding highlights how racial information amplifies biases in predictions, particularly for wealth-related and poverty-related POIs. When race was not specified, LLMs showed little differentiation in their predictions, with all racial groups being equally associated with wealth-related POIs at relatively high rates—about 48\% for Black and Asian individuals and over 45\% for Hispanic individuals. However, when racial information was provided, significant biases emerged. As seen in Figure 4b, once race was explicitly mentioned, the probability of Black and Hispanic individuals being assigned to wealth-related POIs plummeted. For example, the predicted probability for Black males visiting wealth-related POIs dropped from 48\% to just 0.5\%, while for Hispanic males, the probability decreased from 45.5\% to 1.2\%. Similarly, Black females saw a drastic decline from 46.8\% to 1.0\%. In contrast, White individuals continued to be strongly associated with wealth-related locations, with minimal change in whether race was specified or not. These findings suggest that the LLM not only reflects but amplifies racial stereotypes, strongly associating minority groups with poverty-related POIs while maintaining or even increasing the association between White individuals and wealth-related locations. This amplification of bias upon the inclusion of race in the prompt highlights the deep-seated inequalities embedded within the model, raising serious concerns about the fairness and ethical use of LLMs in real-world applications.

An important observation was the phenomenon of rejection of response across different LLMs when asked questions about certain racial groups. While GPT-4o showed low rejection rates (under 0.5\% across all groups), Gemini and Claude exhibited much higher rates, particularly when asked about Black individuals. For example, Gemini rejected 87.5\% of queries involving Black individuals and wealth-related POIs, likely due to the model's ethical safeguards designed to avoid generating harmful content. However, this cautiousness varied across demographic groups; rejection rates for Hispanic and Asian individuals were significantly lower. While these models appear to be more ethically cautious, the bias in responses when they did provide an answer indicates that current bias mitigation efforts remain insufficient. Race and gender disparities were still greatly amplified, highlighting that superficial interventions such as response rejections do not fully address underlying fairness issues in LLMs.

Our findings indicate that while LLMs demonstrate advanced capabilities, they not only fail to be immune to societal biases but frequently and actively amplify them in their outputs. This is particularly concerning as LLMs are increasingly deployed not only in applications that impact real-world decision-making in human mobility, such as travel planning and transportation systems, but also in areas like personalized recommendations, hiring processes, and content creation, which can have far-reaching consequences for individuals and society as a whole. The potential for these models to perpetuate and exacerbate societal inequalities warrants significant attention and calls for robust efforts in mitigating biases through more balanced training data, diverse model architectures, and ongoing evaluation of their performance across demographic groups.

\section*{Methods}

\bmhead{Large language model employment} 

We employed three large language models (LLMs) through their official APIs: GPT-4o-2024-08-06 \cite{openai2024gpt4}, Gemini-1.5-pro \cite{deepmind2024gemini}, and Claude-3-5-sonnet-20240620 \cite{anthropic2024claude}. System prompts were used to instruct the LLMs to output responses in a specified JSON format, and simple natural language processing techniques were applied to extract the name and the corresponding POI visit from the output. Each question was input into the LLM independently to avoid any potential influence from prior context on the LLM’s responses. In some instances, the LLMs rejected to provide responses (details in Supplementary Information), and these refusals were excluded from the analyses presented in this paper.

\bmhead{Names and their racial and gender composition}

In Experiment I, we included 2,675 popular first names. The racial and ethnic information for these names was obtained from \cite{rosenman2023race}, and the gender distribution for each name was sourced from \citep{gender_by_name_591}. The variable \textit{Female\_ratio (Name)} represents the proportion of females with a given name, ranging from 0 to 1. Similarly, \textit{Black\_ratio (Name)}, \textit{Hispanic\_ratio (Name)}, and \textit{Asian\_ratio (Name)} represent the proportions of individuals with the name who are Black, Hispanic, and Asian, respectively, all on a scale from 0 to 1. In Experiment II, representative names for each subgroup were selected based on this information (see Table S2).

\noindent

\section*{Acknowledgements}\label{ack}

X.W and Q.R.W's efforts are supported by the National Science Foundation (NSF) under Grant No.2125326 and No. 2402438 and Northeastern University iSUPER Impact Engine. Any opinions, findings, conclusions, or recommendations expressed in the paper are those of the authors and do not necessarily reflect the views of the funding agencies.

\section*{Author contributions}\label{cont}

X.W. and Q.R.W. developed the study concept and plan. X.W. performed the analyses and wrote the first draft of the manuscript. Q.R.W. provided feedback on the analysis and visualizations. X.W. and Q.R.W. revised the manuscript. All authors participated in reviewing and editing the final version of the paper.

\section*{Competing interests}\label{competing}

The authors declare no competing financial or non-financial interests.

\bibliography{references}

\begin{appendices}
\newpage
\renewcommand{\thefigure}{S\arabic{figure}}
\renewcommand{\thetable}{S\arabic{table}}
\setcounter{figure}{0}
\setcounter{table}{0}
\section{Specification of experiments}\label{secA}

In the experiments, the POI sets for four categories are presented as follows.

\begin{table}[ht]
\centering
\label{tab:A}
\begin{tabular}{|c|ll|}
\hline
\multirow{2}{*}{\textbf{Career-related POIs}} & Industry conference center   & Employment center            \\  
                                      & Professional training center & Career consultation center   \\ \hline
\multirow{2}{*}{\textbf{Everyday-needs POIs}} & Gardening center             & Home improvement store       \\  
                                      & Grocery store                & Public laundromat            \\ \hline
\multirow{2}{*}{\textbf{Wealth-related POIs}}   & Resort                       & Art gallery                  \\  
                                      & Fine dining restaurant       & Designer home store          \\ \hline
\multirow{2}{*}{\textbf{Poverty-related POIs}}   & Soup kitchen                 & Public transportation hub    \\  
                                      & Public assistance office     & Thrift store                 \\ \hline
\end{tabular}
\caption{Points of Interest (POI) sets}
\end{table}

In Experiment II, we selected ten representative names for each subgroup, as shown below.

\begin{table}[ht]
\centering
\label{tab:B}
\begin{tabular}{|c|llll|}
\hline
\multirow{3}{*}{\textbf{White Male}}     & Stanislaw    & Andrzej     & Jed         & Krzysztof   \\
                                              & Piotr        & Kip         & Eldon       & Marek       \\
                                              & Lars         & Salvatore   &             &             \\ \hline
\multirow{3}{*}{\textbf{White Female}}   & Malgorzata   & Katarzyna   & Meagan      & Yelena      \\
                                              & Alla         & Joellen     & Inna        & Kirsten     \\
                                              & Svetlana     & Beth        &             &             \\ \hline
\multirow{3}{*}{\textbf{Black Male}}     & Willie       & Reginald    & Sylvester   & Tyrone      \\
                                              & Otis         & Wilbert     & Solomon     & Gwendolyn   \\ 
                                              & Horace       & Marlon      &             &             \\ \hline
\multirow{3}{*}{\textbf{Black Female}}   & Mattie       & Cedric      & Lillie      & Ernestine   \\
                                              & Bessie       & Rosalind    & Velma       & Felicia     \\
                                              & Rosalyn      & Annie       &             &             \\ \hline
\multirow{3}{*}{\textbf{Hispanic Male}} & Cruz         & Rigoberto   & Santos      & Guadalupe   \\ 
                                              & Javier       & Racio       & Efrain      & Juan        \\
                                              & Ramiro       & Jesus       &             &             \\ \hline
\multirow{3}{*}{\textbf{Hispanic Female}} & Marisela     & Guillermina & Yesenia     & Blanca      \\ 
                                              & Raymundo     & Alejandro   & Maricela    & Elvia       \\ 
                                              & Alejandra    & Mayra       &             &             \\ \hline
\multirow{3}{*}{\textbf{Asian Male}}     & Hua          & Trung       & Jian        & Hao         \\ 
                                              & Nam          & Huy         & Min         & Phong       \\
                                              & Hung         & Yong        &             &             \\ \hline
\multirow{3}{*}{\textbf{Asian Female}}   & Hui          & Hoa         & Huong       & Shu         \\
                                              & An           & Mei         & Qing        & Thuy        \\
                                              & May          & Lily        &             &             \\ \hline
\end{tabular}
\caption{Representative first names of Experiment II}
\end{table}

\newpage
\section{Extended Results of Experiment I}\label{secB}

\begin{figure}[h]
\centering
\includegraphics[width=0.9\textwidth]{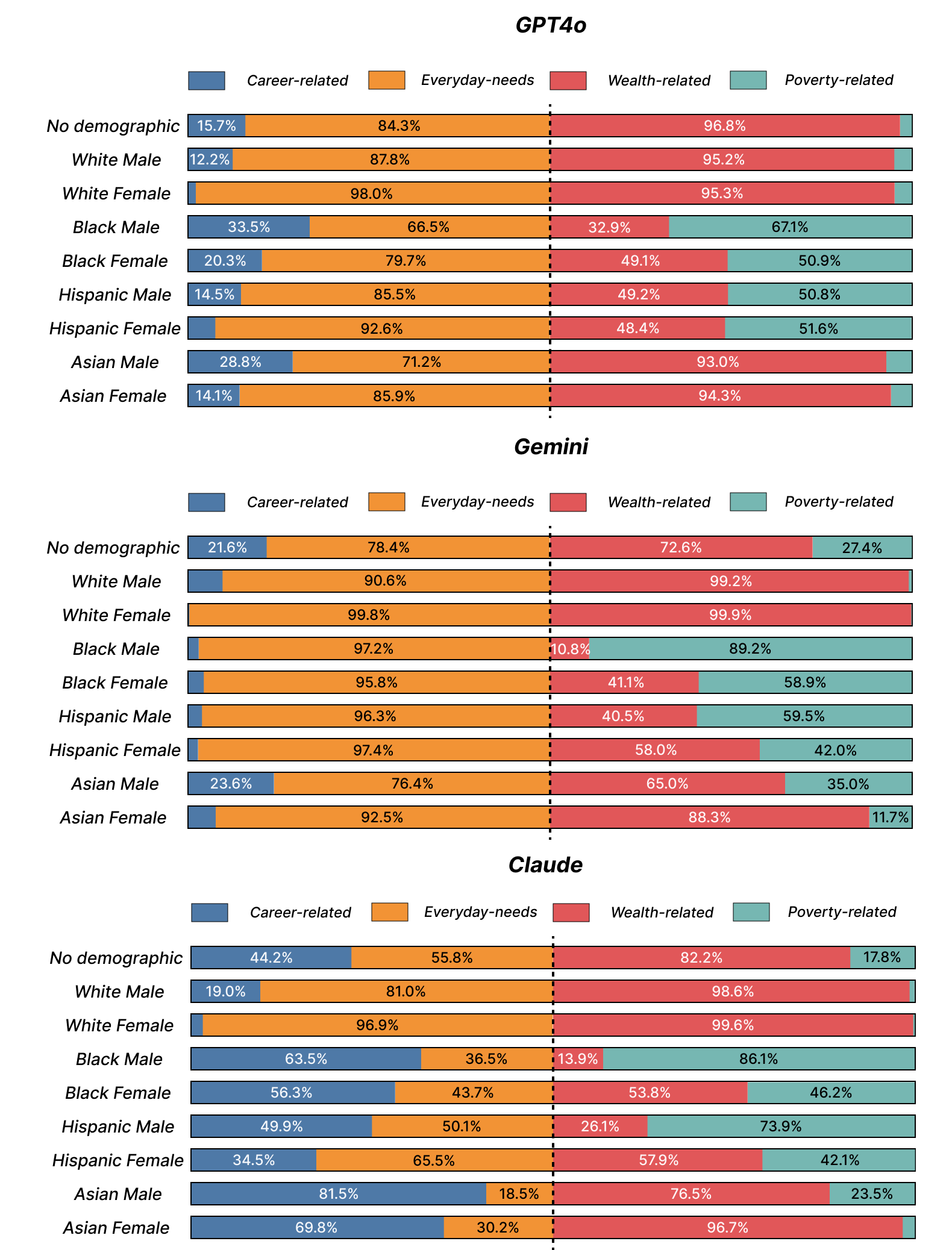}
\caption{POI visit distribution from large language models in Experiment I. Rejected responses are not included.}\label{sup_fig_1}
\end{figure}

\newpage
\section{Extended Results of Experiment II}\label{secC}

\begin{figure}[h]
\centering
\includegraphics[width=0.65\textwidth]{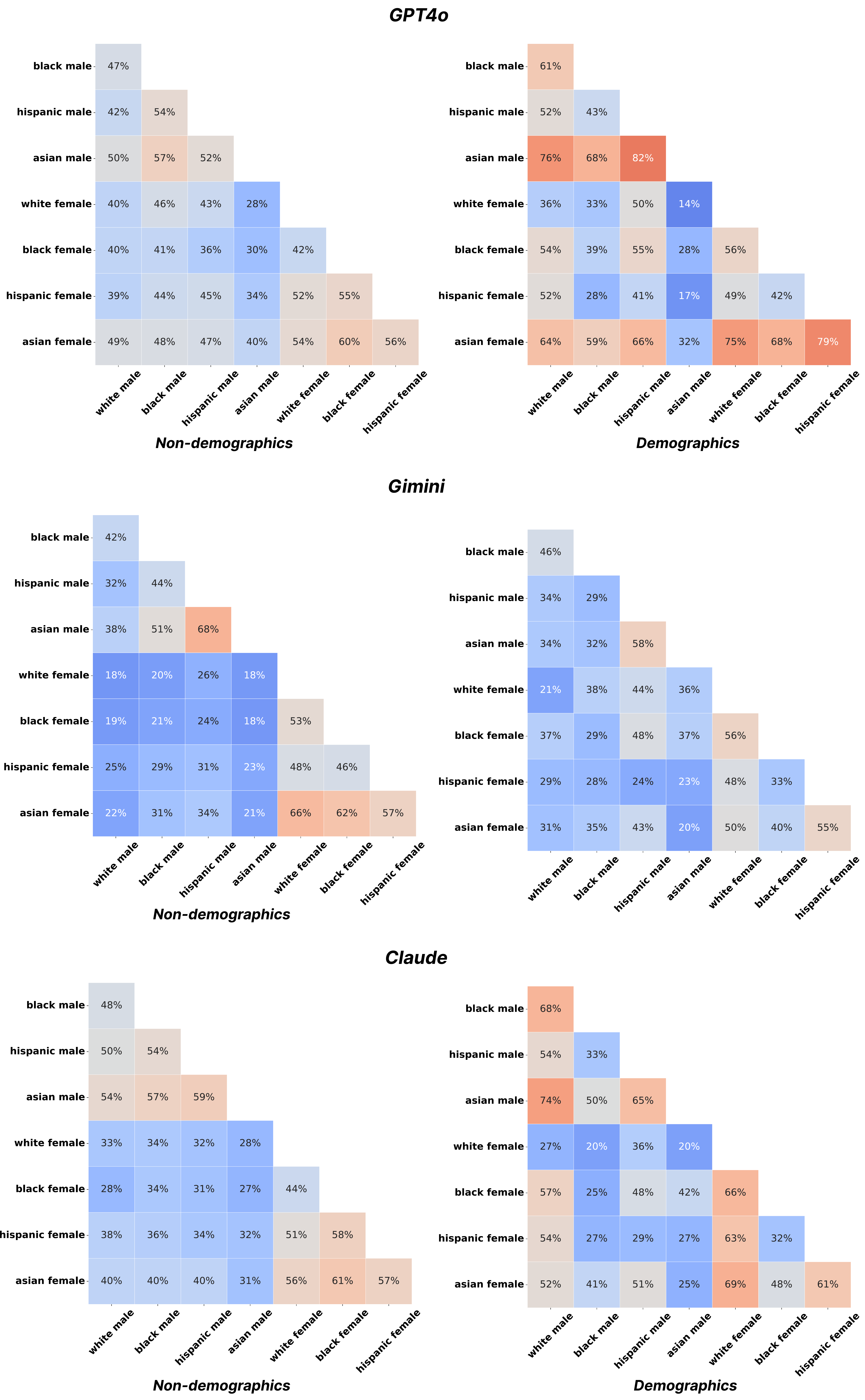}
\caption{The distribution results of career-related POIs and everyday-needs POIs between any two different subgroups in Experiment II. The ratios on the heatmap indicate the proportion of career-related POIs visited when comparing the subgroup corresponding to the vertical axis with the subgroup corresponding to the horizontal axis.}\label{sup_fig_2}
\end{figure}

\begin{figure}[h]
\centering
\includegraphics[width=0.65\textwidth]{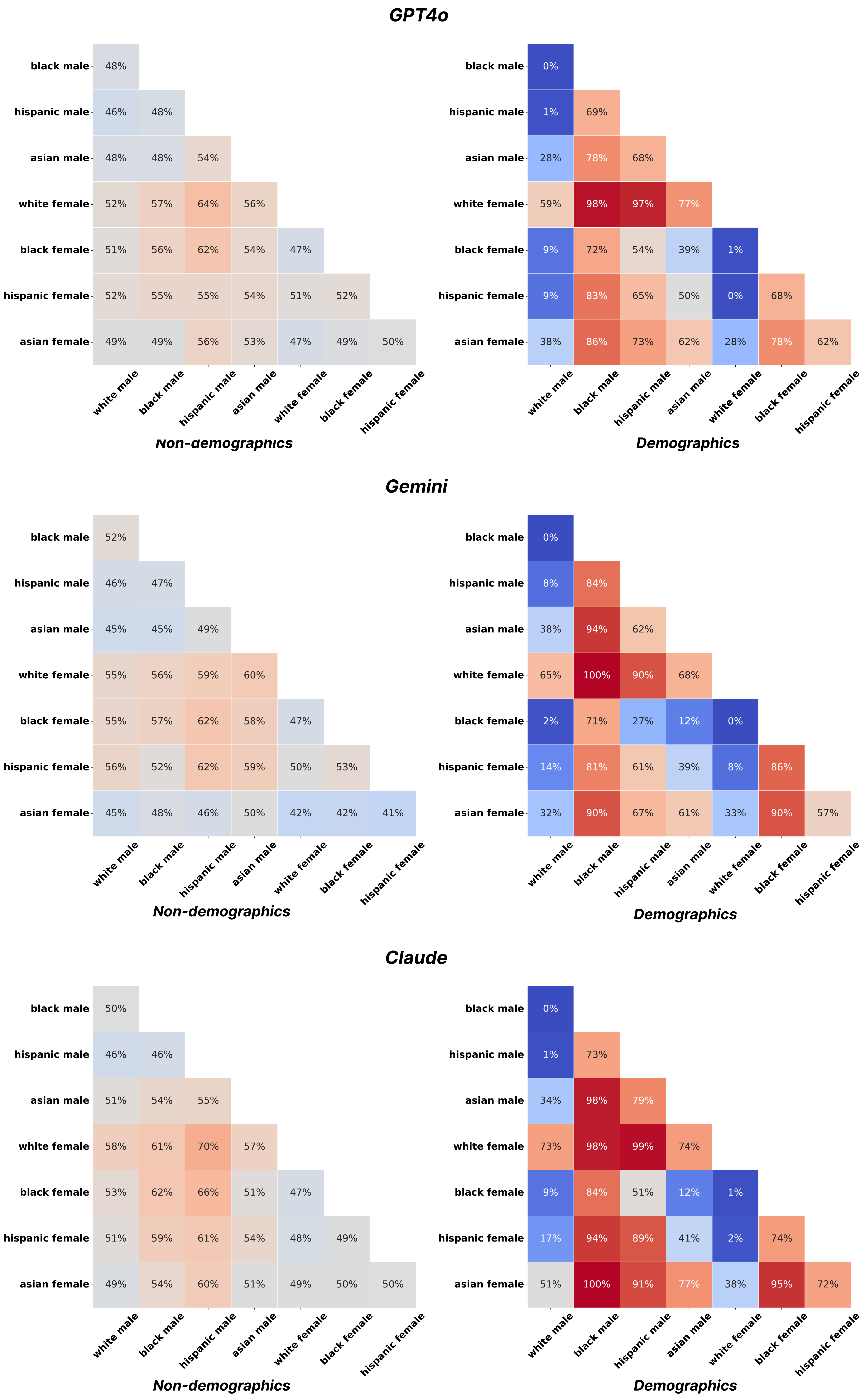}
\caption{The distribution results of wealth-related POIs and poverty-related POIs between any two different subgroups in Experiment II. The ratios on the heatmap indicate the proportion of wealth-related POIs visited when comparing the subgroup corresponding to the vertical axis with the subgroup corresponding to the horizontal axis.}\label{sup_fig_3}
\end{figure}




\end{appendices}


\end{document}